\documentclass[conference]{IEEEtran}
\IEEEoverridecommandlockouts
\usepackage{cite}
\usepackage{caption}
\usepackage{subcaption}
\usepackage{amsmath,amssymb,amsfonts}
\usepackage{algorithmic}
\usepackage[ruled,vlined]{algorithm2e}
\usepackage{graphicx}
\usepackage{textcomp}
\usepackage{xcolor}
\usepackage[para,online,flushleft]{threeparttable}
\usepackage{booktabs,tabulary}
\def\BibTeX{{\rm B\kern-.05em{\sc i\kern-.025em b}\kern-.08em
    T\kern-.1667em\lower.7ex\hbox{E}\kern-.125emX}}

\begin{document}

\title{Robust Neural Regression via %Frustratingly simple 
Uncertainty Learning}
% \title{Towards uncertainty-aware regression networks in metamodelling of large variable annuities portfolio\\}

\author{\IEEEauthorblockN{
Akib Mashrur\IEEEauthorrefmark{1},
Wei Luo\IEEEauthorrefmark{2}, 
Nayyar A. Zaidi\IEEEauthorrefmark{3} and
Antonio Robles-Kelly\IEEEauthorrefmark{4}}
\IEEEauthorblockA{
\textit{School of Information Technology, 
Deakin University,
Waurn Ponds 3216, Australia} \\
Email: \IEEEauthorrefmark{1}amashrur@deakin.edu.au,
\IEEEauthorrefmark{2}wei.luo@deakin.edu.au,
\IEEEauthorrefmark{3}nayyar.zaidi@deakin.edu.au,
\IEEEauthorrefmark{4}antonio.robles-kelly@deakin.edu.au}

ORCID: \IEEEauthorrefmark{1}0000-0002-4404-7471
\IEEEauthorrefmark{2}0000-0002-4711-7543
\IEEEauthorrefmark{3}0000-0003-4024-2517
\IEEEauthorrefmark{4}0000-0002-2465-5971}
% \and
% \and
% \and
% \and
% \and

\maketitle

\begin{abstract}
Deep neural networks tend to underestimate uncertainty and produce overly confident predictions.
Recently proposed solutions, such as MC Dropout and SDENet, require complex training and/or auxiliary out-of-distribution data.
We propose a simple solution by extending the time-tested iterative reweighted least square (IRLS) in generalised linear regression.
We use two sub-networks to parametrise the prediction and uncertainty estimation, enabling easy handling of complex inputs and nonlinear response.
The two sub-networks have shared representations and are trained via two complementary loss functions for the prediction and the uncertainty estimates, with interleaving steps as in a cooperative game. 
% Neural networks implemented in meta-modelling of VA contracts normally optimize the ordinary least squares (OLS) function. However, OLS is not an unbiased estimator when  hetero-skedasticity (non-constant noise) is present in the data. Weighted least square (WLS) loss function could be implemented to achieve unbiased estimates with presence of hetero-skedasticity. However, WLS requires true variance of noise as an weighting parameter, which is not achievable. We propose a two-block neural network with shared layers that can iterativly model conditional variance from model residuals and use the estimated conditional variance to provide unbiased estimate of the mean via optimizing the WLS function. 
Compared with  more complex models such as MC-Dropout or SDE-Net, our proposed network is simpler to implement and more robust (insensitive to varying aleatoric and epistemic uncertainty). 
% We implement the proposed network on a simulated dataset and the benchmark VA dataset. In both cases, our proposed network outperforms the baseline neural network model optimizing ordinary least squares. 
% \wl{toy dataset $->$ simulated dataset}
\end{abstract}

\section{Introduction}
Despite tremendous achievements of Deep Neural Networks (DNNs) in many real life applications, it is known that DNNs tend to make overconfident predictions even in environments with high uncertainty (e.g. high noise in data) \cite{guo2017calibration}. This has the undesired consequence of loss of generalization performance in presence of high uncertainty\cite{wang:2018}. This is because when training sample includes high levels of noise, DNNs often will produce wildly wrong predictions on the validation set \cite{loquercio:2020}. This presents us with the need of uncertainty-aware DNNs that know when high uncertainty is present in the data. Proper quantification of uncertainty and utilizing that uncertainty to get better generalization performance on data with high noise (such as in financial data) can be highly beneficial to many real-life applications \cite{maddox:2019}.

Existing approaches of quantifying uncertainty generally makes prior assumptions on distribution of noise implicitly or explicitly. Moreover, a number of these methods are based on Bayesian Neural Nets (BNNs) \cite{denker1990transforming} or model ensembles \cite{lakshminarayanan2016simple}. Other approaches, such as that in \cite{hafner:2019}  suffer when the task in hand has inherent randomness, i.e. aleatoric uncertainty. For example, in SDEnet \cite{kong2020sde}, to train the \emph{diffusion network}, noise is simulated from a standard Gaussian distribution. Even though, such method can model aleatoric uncertainty properly in a regression environment if noise is normally distributed, the method has two major limitations: (i) Need of auxilary data for modelling noise (this noise is simulated during training) (ii) No direct method of utilizing the modelled noise to achieve unbiased estimate of the mean.  

In contrast, here we take a more data-driven approach to quantifying uncertainty in a regression environment. We propose a multi-task based neural network structure that estimates mean and variance separately from the shared representation of input data. Motivated by the simple \emph{squared residual method} used in statistics for estimating variance, the residuals we achieve from our mean estimation network is applied to fit the variance estimation network. In turn, this estimated variance is used along with the estimated mean to optimize the \emph{weighted least squares}, which is known to provide a more unbiased estimation in presence of hetero-skedasticity (non-constant noise)\cite{li:94}.

Our model provides two benefits over the existing techniques: (1) The model is more data-driven since it directly learns the conditional variance from the residual errors, avoiding the need for stochastic noise simulation using prior assumptions. (2) The estimated conditional variance is directly used to reduce model bias on noisy data, making the model more robust.

To summarize, the main contributions of this paper are:

\begin{itemize}
    \item We provide further evidence for the importance of explicit uncertainty modeling in neural regression models.
    In particular, we show that ignoring uncertainty in regression models leads to poor performance not only in individual predictions but also in the aggregated prediction. 
    And the latter has significant consequence in many applications including portfolio-based finance risk management.
    % point out a major issue in implementing neural networks for real life regression data. Neural networks optimizing a simple least squares loss can provide biased estimates in presence non-constant noise because it violates homoskedasticity assumption of ordinary least squares.
    
    % \wl{OLS has special meaning in statistics. Maybe just say LS; need to explain the abr.}

    \item We develop a neural network implementation of the widely used iterative reweighted least squares (IRLS) procedure, allowing the modeling of nonlinear responses. 
    
    \item We propose a training procedure that involves interleaving the reduction of two loss functions to jointly learn the mean and the variance.
    % that can model inherent uncertainty in regression data.
    
    \item We applied the model and the training procedure on a well-studied risk meta-modeling problem and achieved similar performance some state-of-the-art 
    uncertainty-aware models such as MC-Dropout or SDENet. 
    Note that these are more complex models that are more difficult to train and often requires auxiliary training data.
    % We evaluate the uncertainty-aware regression model on simulated 1D data and the benchmark metamodelling data. Our proposed network achieved much better performance compared to baseline model. 
\end{itemize}

\section{background}

\begin{figure}[!t]
     \centering
     %\begin{subfigure}[b]{0.3\textwidth}
     \begin{subfigure}{0.99\columnwidth}
         \centering
         \includegraphics[width=0.95\columnwidth]{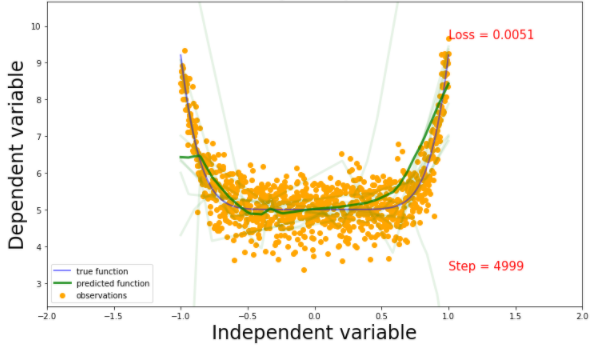}
         \caption{Neural networks can fit data with high homoskedastic (constant variance) noise relatively well even from small training sample}
         \label{fig:nn_homo}
     \end{subfigure}
     %\hfill
     %\begin{subfigure}[b]{0.3\textwidth}
      \begin{subfigure}{0.99\columnwidth}
        \centering
         \includegraphics[width=0.95\columnwidth]{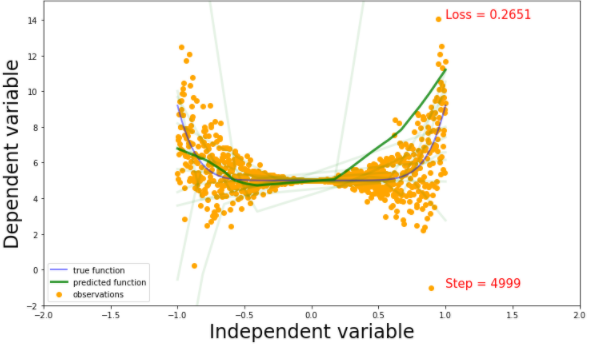}
         \caption{But with heteroskedastic (non-constant variance) noise present in data, a neural network with same structure produces wildly inaccurate results}
         \label{fig:nn_hetero}
     \end{subfigure}
     %\hfill
     %\begin{subfigure}[b]{0.3\textwidth}
     \begin{subfigure}{0.99\columnwidth}
         \centering
         \includegraphics[width=0.95\columnwidth]{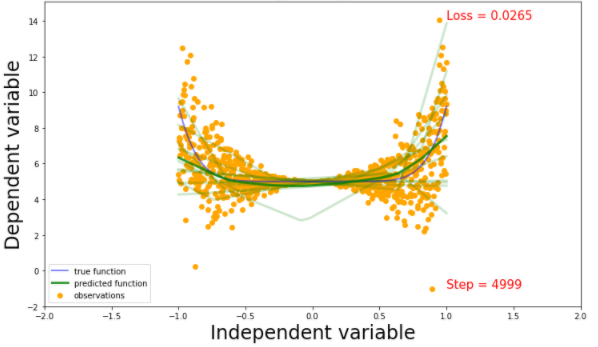}
         \caption{Our proposed model can fit data with heteroskedastic noise much more accurately even from an extremely small training set with same \# of parameters.}
         \label{fig:snn_hetero}
     \end{subfigure}
        \caption{Effects of heteroskedastic noise on performance of neural networks. The solid grey line represents the true function, the solid green line is the predicted regression line (averaged over 10 random experiments represented in light green lines). Only 5\% training data was used for fitting the regression line.}
        \label{fig:nn_noise}
\end{figure}

Capturing uncertainty in data and predictions is crucial for robust neural regression, especially when the training data is small (See Figure \ref{fig:nn_noise}). There already exists significant amount of literature in statistics on dealing with uncertainty and robustness. However, most of these are relevant to simple linear regressions. For example, \emph{Weighted least squares (WLS)}, a generalized version of ordinary least squares can be used to achieve more robust estimates when heteroskedasticity is present in the data. WLS is a principled approach for achieving the best linear unbiased estimation where true conditional variance or noise of data is known \cite{aitken1936iv}. However, the true conditional variance or noise of data (which can be seen as ``Aleatoric uncertainty") is generally intractable in real life. A simple solution called \emph{Iterative re-weighted least square (IRLS)} is a time-tested method to estimate this uncertainty in a linear setting \cite{holland1977robust}. However, the assumption of linearity in such statistical methods restrict their usage to identifying linear dependence only. Real life data can be severely non-linear and high-dimensional, thus limiting the direct application for such simple method.

Uncertainty quantification via neural networks is an active field of research. The current techniques for measuring uncertainty either take a bayesian approach (for example, via imposing prior distributions over model parameters \cite{denker1990transforming}) or ensemble approach (for example, via training multiple DNNs with different random initializations \cite{lakshminarayanan2016simple}). A more recent technique for uncertainty quantification, SDEnet \cite{kong2020sde}, adopts a dynamic systems approach where the diffusion term of the stochastic differential equation (modelled as a neural network) indicates the level of uncertainty inherent in the data. To train this diffusion network, SDEnet simulates noisy data from a Gaussian perturbation of the original input data. This diffusion network uses Brownian motion to encode uncertainty. This indicates the implicit assumption of Gaussian distribution of noise. Another recently popular approach for quantifying uncertainty is the MC-dropout approach \cite{gal2016dropout} which helps to quantify predictive variance with stochastic dropout layers in the model. With dropouts, a binary variable for every parameter in the specified network layer is sampled. The binary variables can take on the value of 1 with a specified probability, also known as the \emph{dropout rate}. The parameter corresponding to the binary variable is dropped if the binary variable takes on the value of 0. To derive uncertainty, multiple stochastic forward passes needs to be used with dropout activated. The results are then averaged to indicate the predictive variance or uncertainty of the model.  This approach is not data driven as the modelled uncertainty can depend on the nature of stochasticity assumed in dropout layers. For example, deactivating a high proportion of nodes in the dropout layer may result in high predictive variance regardless of inherent noise in the data.  MC-Dropout is also known to produce over-confident predictions on unseen examples \cite{lakshminarayanan2016simple}.

% \wl{Cover related work (papers) here. We can always cite our previous IJCNN paper to cover the PE optimisation problem.}

% \wl{Maybe describe problem here. Not existing methods.}

% \subsection{Weighted least squares}

% In statistics, Weighted least square 

% \subsection{Squared residual method}

\section{Uncertainty-aware regression network}
\label{sec:method}

\begin{figure*}[t!]
    \centering
    \includegraphics[width=.98\textwidth]{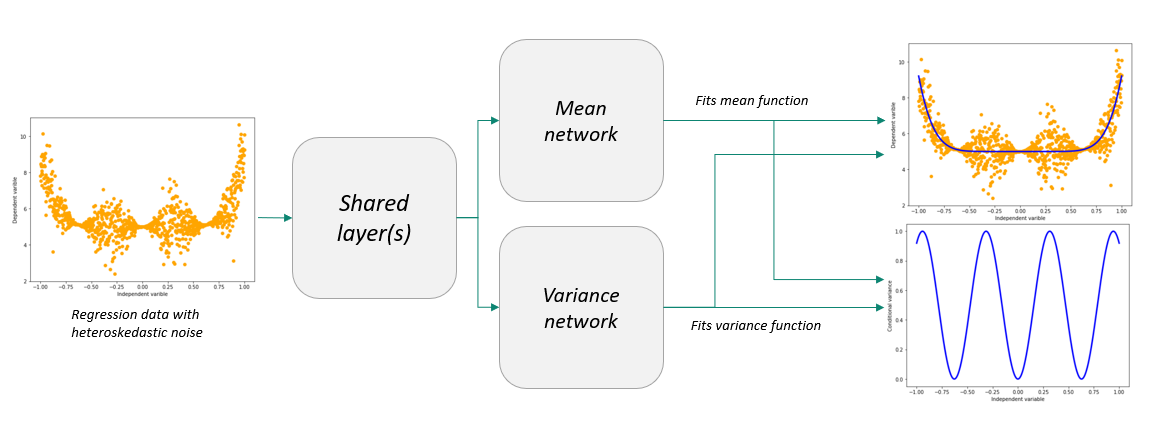}
    \caption{Configuration of the proposed multi-headed uncertainty-aware regression net. Both the \emph{mean network} and \emph{variance network} learns from a shared representation of feature space. The mean networks attempts to fit the true mean function. The variance network attempts to fit the true variance function from the residuals from mean network.}
    \label{fig:iv_network}
\end{figure*}

Inspired by the \emph{IRLS} method for estimating conditional variance, we propose a neural network that employs two different fully connected blocks (mean network and variance network) that attempts to model mean and variance separately from a shared representation of input data. As shown in Algorithm \ref{algo:training}, the squared residuals achieved from the mean network helps to fit the variance network. The fitted mean and variance network is iteratively fitted to optimize weighted least squares, thus giving us an unbiased estimation of mean and also an estimate of squared residuals (variance). The configuration of the proposed network is illustrated in Figure \ref{fig:iv_network}.

\begin{algorithm}[t!]
\SetAlgoLined
% \KwResult{Write here the result }
 Initialize $s,m,v$ \;
 Set $\hat{v}=1$ \;
 \For{\# of epochs}{
  Forward training data through shared network, $s(X)$  \;
  Forward $s(X)$ through the mean network, $m(s(X))$  \;
  Estimate mean, $\hat{y} = m(s(X))$\;
    Calculate squared residuals $X_{res} = {(y-\hat{y})}^2$ \;
Update $s, m$ by $L_m$ \; %$\nabla_{s,m}L_{wls}(\hat{v},y,\hat{y})$
   Estimate variance, $\hat{v} = v(s(X))$\;
   Update $v$ by $L_v$\;%$\nabla_{v}L_{ols}(X_{res},\hat{v})$ 
 }
 \caption{Training of uncertainty-aware regression networks. $s$ is the shared network with fully connected layers, $m$ and $v$ are respectively mean network and variance network,
 $L_m, L_{v}$ are the mean loss and the variance loss as defined in Section~\ref{sec:method}.}
%  $L_{wls}$ is the weighted least square loss, and $L_{ols}$ is the regular mean squred error loss}
 \label{algo:training}
\end{algorithm}

% \wl{This section probably needs more contents. You may define the two loss functions explicitly.}

\subsection{Shared layers}

The shared layers maps the inputs into a continuous manifold, implicitly learning a metric for all inputs. The outputs of this layer is fed in as the input for mean network and variance network.

\subsection{Mean network}

The mean network is similar to the standard deterministic NN regression models.
It, however, has an augmented loss function based on weighted least squares. The weights come from the output of the variance network
% At each training step, the backpropagation will tune the parameters from both the mean network and the variance network.

Given a training batch $\mathcal{B}$, the loss function used to train the mean network and shared layers is:

\begin{equation}
\begin{aligned}
    L_{m} & = \sum_{(X,y)\in \mathcal{B}}\frac{\left(y-\hat{y}\right)^2}{\hat{v}^2} \\
            & = \sum_{(X,y)\in \mathcal{B}}\frac{\left(y-m(s(X))\right)^2}{v(s(X))^2}.
\end{aligned}
\end{equation}

% $$ L_{s,m} = \frac{1}{\hat{v}^2}(y-\hat{y})^2 $$
where $m(s(X)$ is the mean network output and $v(s(X)$ is the variance network output.

Initially, with the untrained variance network, we set uniform $\hat{v}$ so that the loss function behaves like the regular mean squared error. But with the trained variance network at later training steps, we optimize the weighted least squares using the inverse of estimated variances. 

\subsection{Variance network}

The variance network uses the latent embedding from the shared layer and fits a slow changing smooth function for the variance in heteroskedastic data.
The training labels are the residuals, which is determined by the outputs of the mean network.
The loss function used to train this network is defined as:

\begin{equation}
\begin{aligned}
    L_{v} & = \sum_{(X,y)\in \mathcal{B}}\left(\left\|y-\hat{y}\right\| - \hat{v}\right)^2 \\
            & = \sum_{(X,y)\in \mathcal{B}}\left(\left\|y-m(s(X))\right\| - v(s(X))\right)^2.
\end{aligned}
\end{equation}

% $$ L_v = \frac{1}{N}(X_{res}-\hat{v})^2 $$

This can be seen as the mean squared error between the estimated variance and squared residuals from the mean network.

% - At each iteration, Back Propagation will affects the parameters from both the mean network and the variance network.

\subsection{Interleaved training}

Our proposed network follows an interleaved training schedule. For each batch of training data, the model first fits the mean network and shared layers by optimizing $L_{m}$ from provided features and targets. In this stage, since the conditional variances are not estimated yet, we assume an uniform conditional variance for the noise. The parameters of variance networks are not updated during this stage. After fitting the mean network, the squared residuals are then calculated based on mean network's prediction on each batch. The mean network and shared layer parameters are frozen after the initial training and only the variance network is updated to fit the squared residuals by optimizing mean squared loss between the variance network output and calculated squared residuals. The variance network outputs expected squared residuals which is then fed back into the mean network training for optimizing the weighted least square loss, $L_{m}$ for robust predictions.

\section{Empirical evaluation}

To evaluate the performance of our proposed network on scarce and noisy data, we first create a heteroskedastic 1-dimensional simulation data. Then we turn our attention to a real-life data widely used for metamodelling research \cite{gan2018nested}. We chose this dataset because a major challenge in this domain is to achieve a highly robust estimate from an extremely small subset of training samples \cite{luo2020bias}. Since this problem is exactly what we try to solve with our proposed method, we believe that performance in this dataset can be a practical benchmark for real-life applicability of our proposed regression network. 

\subsection{Evaluation of conditional variance estimation}\label{sec:variance_estimation}
To evaluate whether the proposed network can really fit heteroskedastic conditional variance with high accuracy, we simulated a 1D regression dataset $(X,y)$ of 1,000 samples. Our independent variable vector $X$ ranges from -1 to +1,  the mean function applied over X is as below:

\begin{equation}\label{eq:sim_func}
    y = 5+5x^5\sin{(x^3)}+\epsilon.
\end{equation}

$$\epsilon \sim \mathcal{N}(0,v)$$

We can see that the dependent variable $y$ in Equation \ref{eq:sim_func} follows a non-constant Gaussian noise, which has been modelled by:

\begin{equation}\label{eq:var_func}
    v = \frac{5}{2}x^2.
\end{equation}

The conditional variance $v$ in Equation \ref{eq:var_func} determines the scale of noise present in the dependent variable. This can be seen as the inherent scale of noise in the data generation process. Then to replicate the real life challenge of learning noise from extremely scarce set of data, we only keep 1\% of samples for training and 99\% for testing. Over 10 different experiments, we randomly choose different set of training samples (with replacement) and to evaluate overall performance, average the predictions over these 10 random experiments with different training sets of data. 

As shown in Figure \ref{fig:iv_network}, our proposed network contains a shared fully connected layer. The network contains 100 nodes. The output of the layer works as input for both \emph{mean network} and \emph{variance network}. Both mean and variance networks themselves contain 2 hidden layer each activated by the \emph{leaky relu} activation function. The hidden layers for both these networks contain respectively 100 and 50 nodes. The output of the variance network is activated by the \emph{Softplus} activation function to achieve non-negative values for variance estimation.

\begin{figure}[t!]
    \centering
    \includegraphics[width=.98\columnwidth]{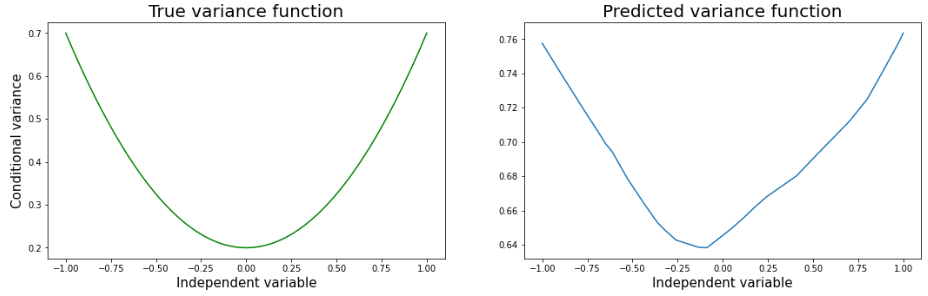}
    \caption{Predicted variance function of our proposed network on simulation data with known true conditional variance. We can see that the network can approximate the heteroskedastic pattern of the conditional variance from a small training sample.}
    \label{fig:variance_func}
\end{figure}

Our simulated 1D data with known true variance of noise allows us to evaluate whether our predicted network can truly capture the conditional variance from data. As shown in Figure \ref{fig:variance_func}, the network can approximate the heteroskedastic characteristics of the conditional variance function.

\begin{figure}[t!]
     \centering
     \begin{subfigure}[b]{0.98\columnwidth}
         \centering
         \includegraphics[width=\textwidth]{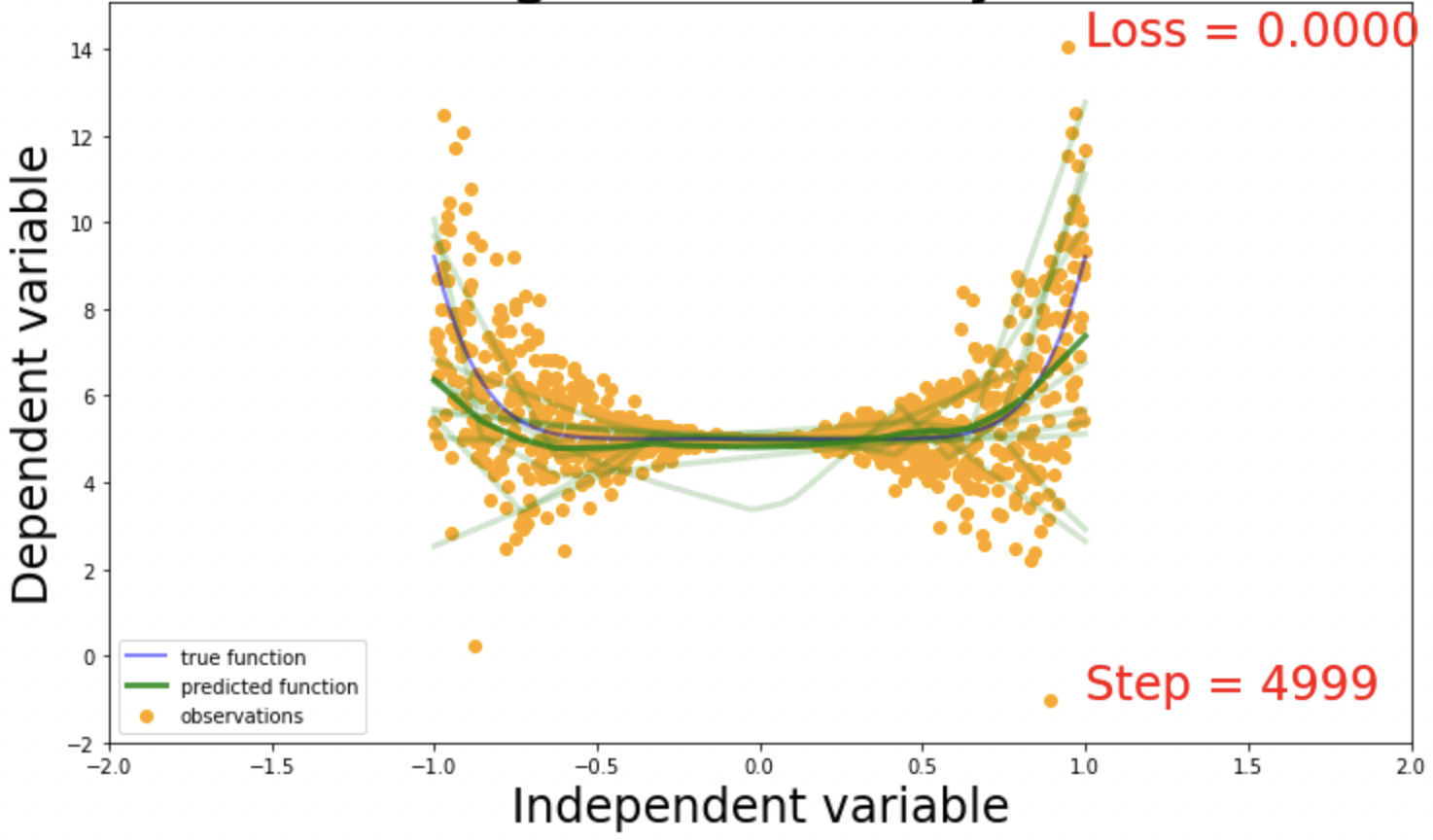}
         \caption{If true sigma was known, a neural network optimizing WLS can fit the heteroskedastic noise properly.}
         \label{fig:nn_homo}
     \end{subfigure}
     \hfill
     \begin{subfigure}[b]{0.98\columnwidth}
         \centering
         \includegraphics[width=\textwidth]{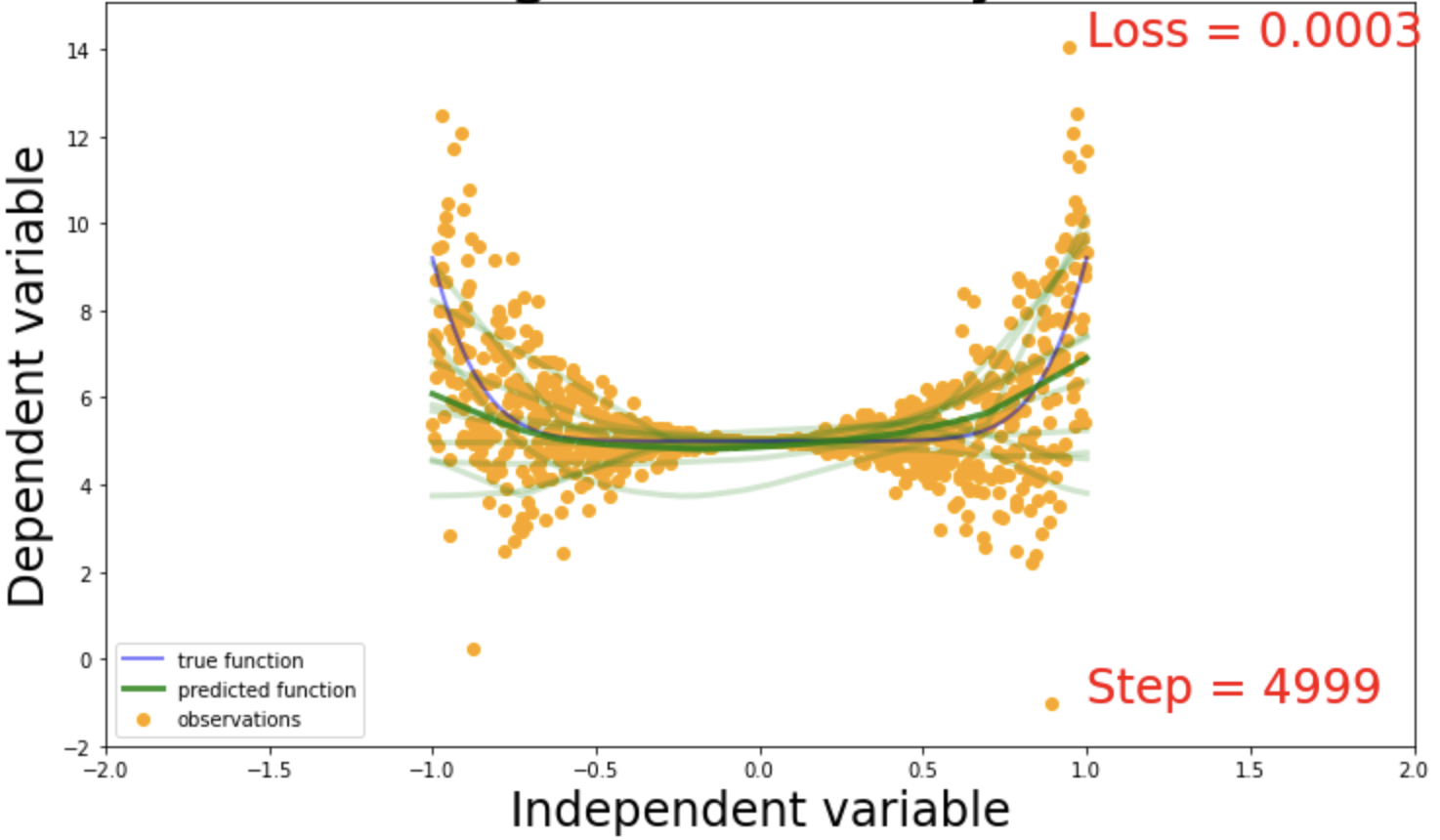}
         \caption{Our proposed model can achieve similar robust estimates with self-estimated conditional variances}
         \label{fig:nn_hetero}
     \end{subfigure}

        \caption{The variance estimated from our uncertainty-aware regression net can give robust estimates similar to a neural network optimizing WLS with knowledge of true variances. The light green lines are predicted regression lines for individual experiments. The darkened green lines is the average of predicted regression lines over 10 experiments. The blue line is the true function.}
        \label{fig:robust_vis}
\end{figure}

\subsection{Evaluation of robust estimation of mean}\label{sec:variance_estimation}

Now that we have established that our network can estimate the variance function with high accuracy, we checked whether the estimated variance can be properly used to get a more robust estimation of the mean. Minimizing the \emph{Weighted least squares} should principally give us the best unbiased estimate of the function under heteroskedastic noise, and we can see that in Figure \ref{fig:robust_vis}. We can observe that with knowledge of true sigma, any neural network could get a robust estimate on our generated data. Since true sigma is not accessible, our proposed network uses the estimated variance to achieve the robust estimates.

\subsection{Performance on benchmark VA data}\label{sec:va_data}

% \begin{table*}[hbt!]
%     \centering
%     \caption{Performance of Inverse Variance Weighting Network on benchmark data-set over 10 experiments}
%     \begin{tabular}{l l l l}
%     \toprule
%         \textbf{Models} & \textbf{No. of Parameters} & \textbf{Avg. PE} & \textbf{Std. of PE}  \\ \midrule
%         Baseline (NN optimizing OLS) & 51.4K & 0.064 & 0.047 \\
%         Proposed model & 51.4K & \textbf{0.055} & \textbf{0.037} \\
%         \bottomrule
%     \end{tabular}

%     \label{tab:iv_va_performance}
% \end{table*}

\begin{table*}[htp!]
  \centering
  \begin{threeparttable}
  \caption{PE Performance on VA dataset across different optimizers}
  \label{tab:performance_flr_va}
  
    \begin{tabulary}{\textwidth}{@{}R RRR C RRR @{}}
         \toprule
       \textbf{Optimizer} & \multicolumn{3}{c}{\textbf{SGDM}}  & \phantom{abc}& \multicolumn{3}{c}{\textbf{Adagrad}} \\
       \cmidrule{2-4} \cmidrule{6-8} 
      & \# of parameters \tnote{\textdagger} & \textbf{Avg. PE} \tnote{*} & Std. of PE \tnote{*} &&  \# of parameters & \textbf{Avg. PE} & Std. of PE \\
      \midrule
      \textbf{Network configuration}\\
      Baseline NN & {51.4K} & {0.072} & {0.041} & & {51.4K} & {0.071} & {0.045}\\
    %   SDEnet& {54k} & {0.090} & {0.077} & & {54K} & {0.951} & {0.171}\\
      MC-Dropout& {51.4k} & {0.073} & {0.041} & & {51.4K} & {0.072} & {0.039}\\
      Uncertainty-aware regression net & {51k} & \textbf{{0.065}} & {0.048} & & {51K} & \textbf{{0.067}} & {0.043}\\
%     \bottomrule
%   \end{tabulary}%
%   \begin{tabulary}{\textwidth}{@{}R RRR C RRR@{}}
    %   \toprule
    \midrule
       \textbf{Optimizer} & \multicolumn{3}{c}{\textbf{RMSprop}}  & \phantom{abc}& \multicolumn{3}{c}{\textbf{Adam}} \\
       \cmidrule{2-4} \cmidrule{6-8} 
      & \# of parameters & \textbf{Avg. PE} & Std. of PE &&  \# of parameters & \textbf{Avg. PE} & Std. of PE \\
      \midrule
      \textbf{Network configuration}\\
      Baseline NN & {51.4K} & {0.069} & {0.038} & & {51.4K} & {0.073} & {0.043}\\
    %   SDEnet& {54K} & {0.794} & {0.092} & & {54K} & {0.751} & {0.080}\\
      MC-Dropout& {51.4K} & {0.072} & {0.054} & & {51.4K} & {0.081} & {0.040}\\
      Uncertainty-aware regression net & {51K} & \textbf{{0.068}} & {0.046} & & {51K} & \textbf{{0.066}} & {0.042}\\
    \bottomrule
  \end{tabulary}
  \begin{tablenotes}
        \item[\textdagger] Total number of learnable parameters in the model.
        
        \item[*] Average and standard deviation of PE across 10 different experiments.
      \end{tablenotes}
    \end{threeparttable}
\end{table*}%

To evaluate the effectiveness of our proposed network in real life application of financial risk modelling, we use a benchmark variable annuity dataset provided by \cite{gan2018nested}. This dataset is widely used by metamodelling researchers with the objective to minimize the percentage error (PE) in portfolio-level risk estimation with limited available sample due to expensive financial simulation procedures.

A detailed description of the dataset is given in \cite{gan2018nested}. Even though the response variable in the dataset is the monthly Delta (portfolio risk measure) over a period of 30 years, for simplicity, we only aim to use the network for estimating the first month's Delta. 

Generally in metamodelling settings, a clustering (generally K-prototype clustering) is implemented first to identify the most representative contracts to reduce the financial simulation process. Then a regression method (generally \emph{Krigging}), is applied on the representative contracts. The fitted regression model is then used to estimate the full portfolio-level risk measure. However, in such cases, the performance of the initial clustering process can highly determine the performance of the final regression network since the clustering process needs to ensure that the data sampled is most representative of the full portfolio. Thus, the accuracy of clustering strongly affects the out-of-sample performance of any regression network trained on those samples. This process can be seen as indirectly eliminating non-representative or noisy data from training set to enhance the generalization performance of regular regression networks.

To eliminate such dependence of clustering techniques from our experiments, we use a random sampling method instead, only sampling 1\% of training data (380 contracts) to estimate Delta of the 38,000-contracts VA portfolio.

To adapt to this far more difficult challenge of robust estimation on the benchmark VA data from 1\% training sample, we use a relatively larger network than previously used for simulation data. For our experiments on this benchmark VA data, we increased the number of nodes in our shared layer to 200. The mean and variance network consists of 200 nodes in each of their 2 hidden layers. We also use a cyclical learning rate schedule with maximum learning rate of 0.01 and minimum learning rate of 0.001 and 100 steps per cycle.

Also, to ensure that our proposed network can be generalised across different types of optimisers, we used some commonly used optimizers (SGD with momentum, AdaGrad, RMSProp and Adam) for our evaluation.

For metamodelling research, since the main business objective is to measure portfolio-level risk measure with high accuracy, percentage error or PE is used as the most common metric for evaluation of meta-models. The PE can be formulated as: 

\begin{equation}
\mathrm{PE} = \frac{\sum_{c_{i} \in \mathcal{P}} \hat{y}_i - \sum_{c_{i} \in \mathcal{P}} y_i}{\sum_{c_{i} \in \mathcal{P}} y_i}.
\end{equation}

We used this metric to evaluate our model against the baseline models. To get a better understanding of the consistency of our model performance, we calculated the average PE and standard deviation of PE over 10 random experiments on the VA dataset using different training samples.

Table \ref{tab:performance_flr_va} shows the results of our model against the baseline models (a fully connected neural network optimizing mean squared error and MC-dropout with similar number of parameters) across all different optimizers with a fixed learning rate. We can see that across all these different optimizers, our proposed model consistently provides similar performance compared to the baseline model with simpler model architecture and training schedule.
% Table \ref{tab:performance_clr_va} shows these results when a cyclical learning rate was used. 

\section{Conclusion}

In this paper, we proposed a novel approach towards estimating uncertainty in a regression data. Our proposed approach also integrates the estimated uncertainty to provide more robust estimations from extremely scarce and noisy data. Our proposed method estimates conditional variance in data from fitting the squared residuals from the mean network. The network then optimizes the weighted least squares loss utilising the inverse of estimated variances as the weights for minimizing the squared loss. Our network is able to fit the mean and conditional variance function separately from the latent representation of input data. We evaluated this network on our simulated data with heteroskedastic noise and the benchmark variable annuity dataset. After evaluating our network with different widely used network, we can conclude that our network can provide more robust estimates even from extremely small training samples. 

To our knowledge, such direct method of estimating uncertainty in regression data from squared residuals via neural network has not been applied before. This extremely simple and intuitive method can highly benefit real-life application domains, such as in metamodelling where robust aggregate estimation is crucial from extremely small and noisy training samples. Naturally, our next step is to develop the network to model aleatoric uncertainty and epistemic uncertainty separately. This would allow us to apply this robust network in recent active learning setups with acquisition rules that consider both epistemic and aleatoric uncertainty \cite{hafner2020noise}. In future we also want to integrate the loss function proposed in \cite{luo2020bias} with our proposed network for faster unbiased estimation of aggregate-level portfolio measures.

\bibliographystyle{IEEEtran}
\bibliography{bibliography.bib}

\end{document}